\documentclass{article}
\usepackage{arxiv}
\usepackage[utf8]{inputenc} 
\usepackage[T1]{fontenc}
\usepackage[english]{babel}
\usepackage{hyperref}       
\usepackage{url}     
\usepackage{booktabs}     
\usepackage{amsfonts}     
\usepackage{nicefrac}    
\usepackage{microtype}      
\usepackage{booktabs}       
\usepackage{amsfonts}       
\usepackage{nicefrac}       
\usepackage{microtype}      
\usepackage[square,numbers]{natbib}
\usepackage{amsmath}
\usepackage{graphicx}
\graphicspath{{images/}}
\usepackage{subfigure}
\usepackage{array}
\usepackage{url}
\usepackage{float}
\usepackage{multicol}
\usepackage{makecell}
\usepackage{colortbl}
\usepackage{amssymb}
\usepackage{tikz}
\usepackage{xcolor}
\usepackage{mathtools}
\usepackage{rotating}
\usepackage{xcolor,colortbl}
\definecolor{Pink}{rgb}{1, 0.7, 0.8}
\definecolor{Green}{rgb}{0,0.5,0}
\definecolor{LightBlue}{rgb}{0.6,0.8, 0.9}
\definecolor{Yellow}{rgb}{1,1,0}
\definecolor{SkyBlue}{rgb}{0.5,0.8,0.9}
\usepackage{gensymb}
\usepackage[pages=some]{background}
\usetikzlibrary{calc,trees,positioning,arrows,chains,shapes.geometric,%
    decorations.pathreplacing,decorations.pathmorphing,shapes,%
    matrix,shapes.symbols}
\tikzset{
>=stealth',
  punktchain/.style={
    rectangle, 
    rounded corners, 
    fill=blue!30,
    draw=black, very thick,
    text width=8em, 
    minimum height=3em, 
    text centered, 
    on chain},
  line/.style={draw, thick, <-},
  element/.style={
    tape,
    top color=white,
    bottom color=blue!50!black!10!,
    minimum width=8em,
    draw=blue!40!black!10, very thick,
    text width=10em, 
    minimum height=3.5em, 
    text centered, 
    on chain},
  every join/.style={->, thick,shorten >=1pt},
  decoration={brace},
  tuborg/.style={decorate},
  tubnode/.style={midway, right=2pt},
}

\title{A psychophysiologically oriented \\ saliency map prediction model}
\author{
 Qiang Li \\
  Image Processing Laboratory\\
  University of Valencia\\
  Valencia,  46980, Spain \\
  \texttt{qiang.li@uv.es} \\
}

\begin{document}
\maketitle
\begin{abstract} \normalsize

\textcolor{black}{Visual attention is one of the most significant characteristics for selecting and understanding the outside redundancy world. The human vision system cannot process all information simultaneously due to the visual information bottleneck. In order to reduce the redundant input of visual information, the human visual system mainly focuses on dominant parts of scenes. This is commonly known as visual saliency map prediction. This paper proposed a new psychophysical saliency prediction architecture, WECSF, inspired by \textit{multi-channel model} of visual cortex functioning in humans. The model consists of opponent color channels, wavelet transform, wavelet energy map, and contrast sensitivity function for extracting low-level image features and providing a maximum approximation to the human visual system. The proposed model is evaluated using several datasets, including the MIT1003, MIT300, TORONTO, SID4VAM, and UCF Sports datasets. We also quantitatively and qualitatively compare the saliency prediction performance with that of other state-of-the-art models. Our model achieved strongly stable and better performance with different metrics on natural images, psychophysical synthetic images and dynamic videos. Additionally, we found that Fourier and spectral-inspired saliency prediction models outperformed other state-of-the-art non-neural network and even deep neural network models on psychophysical synthetic images. It can be explained and supported by the \textit{Fourier Vision Hypothesis}. In the meantime, we suggest that deep neural networks need specific architectures and goals to be able to predict salient performance on psychophysical synthetic images better and more reliably. Finally, the proposed model could be used as a computational model of primate vision system and help us understand mechanism of primate vision system. The project page can be available at: \url{https://sinodanishspain.github.io/HVS_SaliencyModel/}.
}   
\end{abstract}

\keywords{Visual attention \and Redundancy \and Multi-channel model \and Opponent color channel \and Wavelet energy map \and Contrast sensitivity function \and Saliency prediction}

\section{Introduction}

The human vision system (HVS) uses visual attention to extract information from the redundancy of the natural world~\citep{Treisman80}. Redundancy in natural scene images is generally ineffective for scene classification or recognition. \textcolor{black}{Focused visual attention can be used to identify and remove irrelevant information from a cluttered natural environment, according to Barlow's \textit{efficiency-coding hypothesis}~\citep{Barlow59}. Bottom-up and top-down visual attention mechanisms are the most common. Models built from the bottom up are primarily motivated by external stimuli. A variety of visual information (such as color, frequency, texture, orientation, and motion) is processed to extract image features using ~\citep{Wang05}. Contrast this with bottom-up approaches that aim to achieve a specific goal, which typically involve high-level information feedback and modulate lower-level vision functions~\citep{Itti20}. Saliency map prediction has been successfully applied to both of the above computational neural models, and long-term research has been done on the visual attention mechanism~\citep{Sun03}}.
 
Many studies have tried to carry out bottom-up or top-down computational modeling to predict saliency maps. In the following, I will briefly review some saliency prediction models that have achieved remarkable performance in saliency prediction. One of the earliest computational models, proposed by Itti et al.~\citep{Itti98}, was based on the bottom-up mechanisms of low-level vision systems. The model structure contains linear filtering, center-surround differences, across-scale combinations, and linear combinations. Achanta~\citep{LCAV-CONF-2009-012} devised a model for area segmentation that generates saliency maps to identify standout objects with clearly defined boundaries. Bruce and Tsotsos~\citep{Bruce05} proposed a model based on Shannon's self-information assessment to identity saliency map. Based on an investigation of the amplitude spectrum of natural images, Li~\citep{Jian12} presented a novel bottom-up computational model for determining visual saliency. Hou and Zhang~\citep{Hou08} proposed a spatial-temporal visual attention model based on feature rarity. Using the incremental coding length (ICL) method, they figured out each feature's perspective entropy gain and then made a saliency map. Murray et al.~\citep{Murray11} showed a color appearance model could be used for producing saliency maps because it involves parameter selection and spatial pooling function. Boolean map-based saliency (BMS) is a novel Boolean map-based saliency model created by Zhang and Sclaroff~\citep{Zhang13}. According to Gestalt's theory of \textit{figure-ground segregation}, the BMS model computes saliency maps by examining Boolean maps' topological structure~\citep{Baingio17}. A simple image descriptor known as the image signature was established by Hou et al.~\citep{Hou11}. They developed a saliency algorithm based on the image signature. Goferman~\citep{Goferman10} described a context-aware saliency approach that seeks to recognize image regions that represent the scene. According to this approach, instead of identifying fixation locations, the dominating item is detected. Hou and Zhang~\citep{Hou07} demonstrated a straightforward approach for generating the corresponding saliency map in the spatial domain by analyzing the log-spectrum of an input image. Guo~\citep{Guo08} proposed a fast method that uses the spectral residual of the amplitude spectrum to build the saliency map. Schauerte and Stiefelhagen~\citep{Schauerte12} firstly proposed employing Eigenaxes and Eigenangles for models of spectral saliency dependent on the Fourier transform. Murray~\citep{Murray13} proposed a saliency model based on a low-level spatial-chromatic function of HVS, which successfully predicted chromatic induction phenomena and generated a saliency map~\citep{Otazu10}. To detect both static and spatial-temporal saliency, Seo and Milanfar~\citep{Seo09} provided an innovative, unifying computational framework for detecting saliency. Wavelet transforms have also begun to be widely used to estimate computational vision saliency maps~\citep{Murray11}. Compared to the Fourier transform, it has a high resolution in both the frequency and time domains. Wavelet can decompose signals at different scales, also known as multi-resolution/multi-scale analysis or sub-band coding, which can capture more low-level information from the original signal~\citep{Louis97}. On the other hand, the wavelet transform approach can explain the primary visual cortex (V1) properties and produce multi-scale and multi-orientation features when provided with stimuli. The final estimated saliency map could sum up all the processed wavelet coefficients through an inverse wavelet transform~\citep{Selvaraj09}. When using the wavelet transform to create saliency maps, however, global contrast is lost rather than local information. Spratling~\citep{Spratling11} proposed a saliency prediction model based on \textit{predictive coding theory}~\citep{Spratling16}.

In the last few years, some studies have attempted to estimate saliency maps with deep convolutional neural networks (CNNs), which have achieved impressive performances compared to conventional methods~\citep{Borji19, Kruthiventi16}. Cornia et al.~\citep{Cornia16} introduced a novel deep architecture for saliency prediction. To predict saliency maps, current state-of-the-art models for saliency prediction use fully convolution networks, which utilize a non-linear combination of features extracted from the last convolution layer. DeepGazeII~\citep{Kummerer16} is a model which forecasts where people will look in images. The model employs features from the VGG-19 deep neural network, which has been trained to recognize objects in images. Compared to conventional methods, deep learning implementations are also easier to transfer to real-life applications such as object detection, video understanding, and image compression. In this study, our main contributions are fourfold:

\begin{enumerate}
\item A psychophysically oriented saliency prediction model was proposed, which was inspired by the \textit{multi-channel model} of human vision system function. The model has a contrast sensitivity function, an opponent color channel, a wavelet transform, a wavelet energy map, and a wavelet transform energy map. The proposed model is a bottom-up model, and it was tested on the different datasets using certain metrics. 

\item The spatial chromatic contrast sensitivity function was implemented by Python\footnote{https://www.python.org/}, which is available at \url{https://github.com/sinodanishspain/CSFpy}.

\item \textcolor{black}{The proposed model achieved strongly stable and better performance with different metrics on natural images, psychophysical synthetic images, and dynamic scenes. Beyond the accuracy of saliency prediction, we take more neuroscience concepts into account rather than statistical concepts, and it's also another computational goal in the study.}

\item We found that Fourier and spectral-inspired saliency prediction models outperformed other state-of-the-art non-neural network and even deep neural network models on psychophysical synthetic images. It can be explained and supported by the \textit{Fourier Vision Hypothesis}. The proposed model can be successfully applied to explain the "pop-out" effects in the visual search and attention mechanisms of primate vision systems and inspire the development of better deep learning models. \textcolor{black}{Furthermore, we suggest that deep neural networks require distinct architectures and goals in order to reliably predict salient performance on psychophysical synthetic images.}  

\end{enumerate}

The rest of this paper is organized as follows: Section 2
~introduces the concepts of opponent color space, wavelet decomposition, wavelet energy map estimation, and CSF. Section 3
~introduces the saliency map prediction model, along
~with different datasets and evaluation metrics. Section 4 presents the experimental results. The final section provides discussions and conclusions for the paper.

\section{The Proposed Saliency Prediction Model}

\subsection{Saliency Prediction Model}

\begin{figure}[H]
    \centering
    \includegraphics[width=\textwidth]{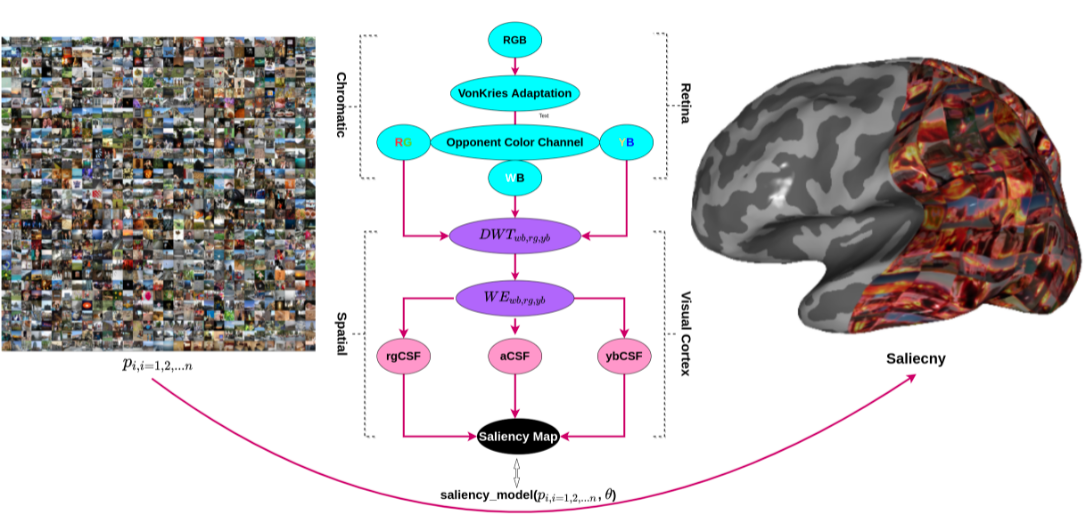}
    \caption{\textbf{Architecture of the proposed saliency prediction model.} The left panel image was selected from the MIT1003 dataset. The flow chart shows the framework of the proposed model, containing the chromatic response in the retina and spatial feature processing in the visual cortex. The natural
~image is first adapted, before decomposing it into white-
black, red-green, and yellow-blue opponent neural channels. In the spatial component, a discrete wavelet transform is applied to each opponent color channel, then the wavelet energy map is measured. In the last step of the proposed model, the CSF is applied to each opponent wavelet energy channel and combined with each opponent's feature. $i$ and $\theta$ indicate image and model parameters,
 respectively. The details of each component are described in the following section. The graph~on the right refers to the map of the left panel image's saliency on the inflated visual cortex using the proposed model.}
    \label{Fig.model_arch}
\end{figure}

In this paper, we propose a biologically inspired visual saliency prediction map, based on the human low-level visual system. The extraction of information from the retina, LGN, and V1 is a critical component of visual neural networks. The color opponent channel, wavelet transform, wavelet energy map, and contrast sensitivity function are the main components of the proposed model architecture. The color opponent channel simulates the response of retinal cells to different spectral wavelengths, and the wavelet transform presents the multi-scale and multi-orientation properties of the V1. The CSF is used to describe the human brain's susceptibility to spatial frequencies. The details of each component are described in the following sections. Fig.~\ref{Fig.model_arch} depicts the computational saliency prediction model architecture. 

\subsection{Gain control with von Kries chromatic adaptation model}

\textcolor{black}{Gain control exists in the visual information processing pipeline in the retina and cortex. In other words, gain control influences both top-down and bottom-up visual information flows, as well as attention-related cognitive functioning~\citep{Butz04}.} Meanwhile, gain control always strives to maintain a steady-state brain and self-regulation condition between the brain and the natural environment. In the von Kries model, we multiply each channel of the image with the gain value after normalizing its intensity~\citep{Finlayson93, Finlayson02, Krauskopf92}. However, there are some implications to this approach. The first is that the channels are considered independent signals, which is why we use independent gains. Second, this gain is added not in the RGB space but, instead, in the tristimulus LMS space. Assuming that the LMS is the same as our image's tristimulus values, the von Kries model can be written in math as: 

\begin{equation}
L_{2}=\frac{L_{1}}{L_{\max }} L_{\max 2},
M_{2}=\frac{M_{1}}{M \max } M_{\max 2},  
S_{2}=\frac{S_{1}}{S \max}  S_{\max 2},
\end{equation}

\begin{equation}
\left[\begin{array}{c}
L_{post} \\
M_{post} \\
S_{post}
\end{array}\right]=\left[\begin{array}{ccc}
\frac{1}{L_{\max }} & 0 & 0 \\
0 & \frac{1}{M_{\max }} & 0 \\
0 & 0 & \frac{1}{\sin a x}
\end{array}\right]\left[\begin{array}{c}
L_{1} \\
M_{1} \\
S_{2}
\end{array}\right],
\end{equation}

\begin{equation}
\left[\begin{array}{c}
L_{2} \\
M_{2} \\
S_{2}
\end{array}\right]=\left[\begin{array}{ccc}
L_{\max 2} & 0 & 0 \\
0 & M_{\max 2} & 0 \\
0 & 0 & S_{\max 2}
\end{array}\right],
\end{equation}

where $L_{1}$ corresponds to the original image's
~L
~values; $L_{Max}, M_{Max}$, and $S_{Max}$, respectively, correspond to the maximum value of each channel in the LMS image; $L_{Max2}, M_{Max2}$, and $S_{Max2} $ are the gain values with a set value of 0.6 in the proposed model; and $L_{2}$ is the corrected L
~channel after adaptation. 

\subsection{Color Appearance Model}

\textcolor{black}{Representation of color in the brain can improve object recognition and identity. Trichromatic theory~\citep{Brill14} and the color appearance model proposed based on the functioning of the sensors encode color information and have been widely used in low-level image processing. Two functional types of chromatic sensitivity or selectivity sensors were found---single-opponent and double-opponent neurons---based on the responses of long (L), mediate (M), and short (S) cones in the physical world~\citep{Shapley11}. Most saliency prediction models use CIELAB and YUV color spaces for the opponent color spaces. In our case, we use another opponent's color space~\citep{Hering1920, Hurvich57}, and the color space transform matrix from RGB to $O_{1}O_{2}O_{3}$ can be expressed as:} 

\begin{equation}
\left[\begin{array}{l}
O_{1} \\
O_{2} \\
O_{3}
\end{array}\right]=\left[\begin{array}{ccc}
0.2814 & 0.6938 & 0.0638 \\
-0.0971 & 0.1458 & -0.0250 \\
-0.0930 & -0.2529 & 0.4665
\end{array}\right]\left[\begin{array}{l}
R \\
G \\
B
\end{array}\right].
\end{equation}

\begin{figure}[H]
    \centering
    \includegraphics[width=\textwidth,height=13.3em]{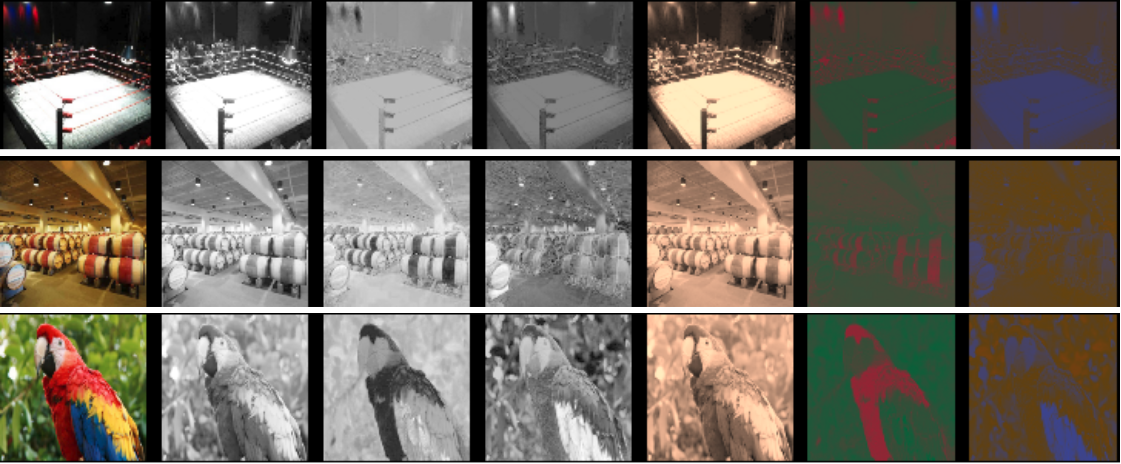}
    \caption{\textbf{Opponent color processing.} The first column represents the raw RGB color space, followed by the white--black (WB) channel, red--green (RG) channel, and yellow--blue (YB) channel, each with a gray colormap. The final three columns depict the WB, RG, and YB channels in artificial color, in order to better visualize the opponent color processing in the visual system.}
    \label{Fig.ATD_IMGS}
\end{figure}

The test natural%
~scene images (of sizes $256\times256$ and $512\times512$
) were selected from the Signal and Image Processing Institute, University of Southern California \footnote{\url{http://sipi.usc.edu/database/database.php?volume=misc}}, and the Kodak lossless true-color image database \footnote{\url{http://r0k.us/graphics/kodak/}}(of sizes $512\times768$, $768\times512$, and $768\times512$
). The total natural%
~color images were resized into the same size (8 bits, $256\times256$%
) as test images. All natural%
~chromatic images were converted from RGB space to the $O_{1}O_{2}O_{3}$ domain, based on the above conversion matrix. As can be seen in Fig.~\ref{Fig.ATD_IMGS}, the chromatic information (white-black, red-green, and yellow-blue) was decomposed into each channel.

\subsection{Wavelet Energy Map}

\subsubsection{Visual cortex receptive fields with wavelet filters}

\begin{figure}[!ht]
    \centering
    \includegraphics[width=\textwidth, height=26em]{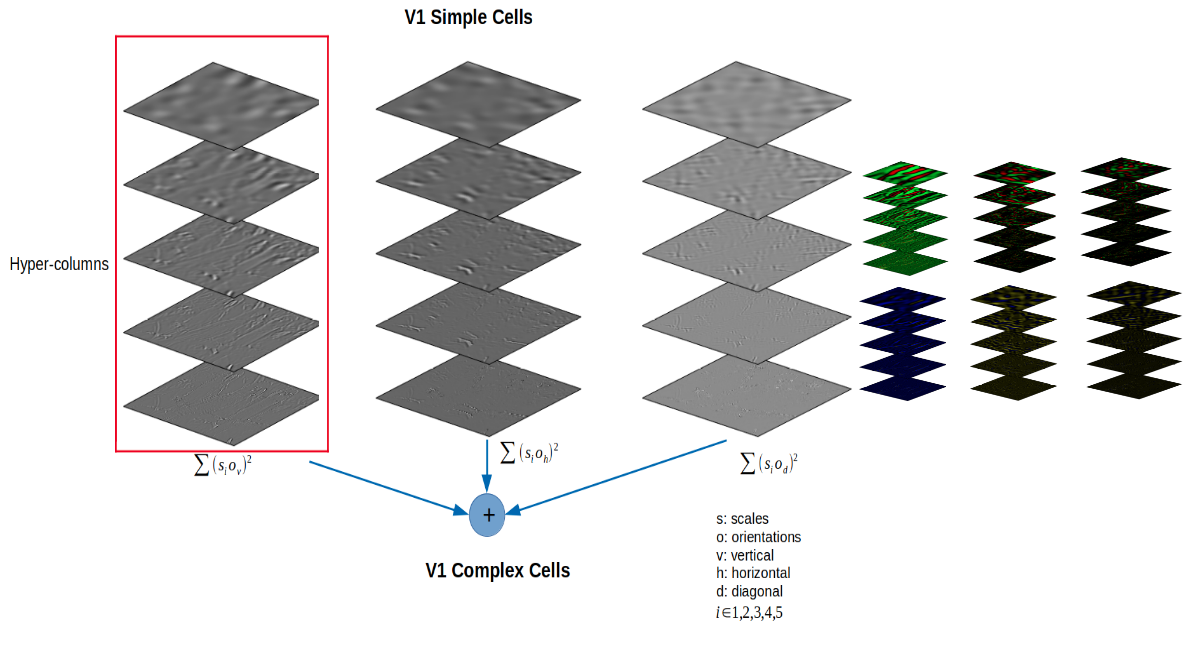}
    \caption{\textbf{The modeling of V1 simple and complex cells in each opponent channel.} The red rectangle indicates hypercolumns
~in the visual cortex. From left to right, the graph depicts WB opponent neurons with different orientations and scales. The following zoomed out top/bottom graphs with artificial color, for better visualization of features, in each hypercolumn indicate RG/YB opponent neurons across different orientations and scales. The V1 complex cells can be obtained from the sum of squares of
~wavelet transform features across scales and orientations in the simple cells.}
    \label{Fig.V1}
\end{figure}

The primary visual cortex contains neurons that reflect the structure of the retinal image in terms of a wavelet basis, and the visual simple and complex cells can be modeled with wavelet filters. In our case, we did not consider the in-depth details of each hypercolumn neuron's interaction
~mechanisms (e.g., Li's model~\citep{Zhaoping98}). The simulated V1 complex receptive fields sum all the squares of different scales and orientations after the wavelet transform (see Fig. \ref{Fig.V1}). The V1 simple receptive fields in each opponent channel are mathematically defined as:

\begin{align}
V_{iv} = s_{i}o_{v} \\  V_{ih} = s_{i}o_{h} \\ V_{id} = s_{i}o_{d},
\end{align}

where $s$ indicates receptive filed scales, $o$ refers to orientation---that is, vertical ($v$), horizontal ($h$), and diagonal ($d$)---and $i$ indicates the number of neurons/features. The V1 complex cells can be formulated as:

\begin{equation}
    V_{complex} = \sum_{1}^{i} (s_{i}o_{v})^2 + \sum_{1}^{i} (s_{i}o_{h})^2 + \sum_{1}^{i} (s_{i}o_{d})^2 .
\end{equation}

\subsubsection{Wavelet transform and wavelet energy map}

\begin{figure}[!ht]
    \centering
    \includegraphics[width=\textwidth]{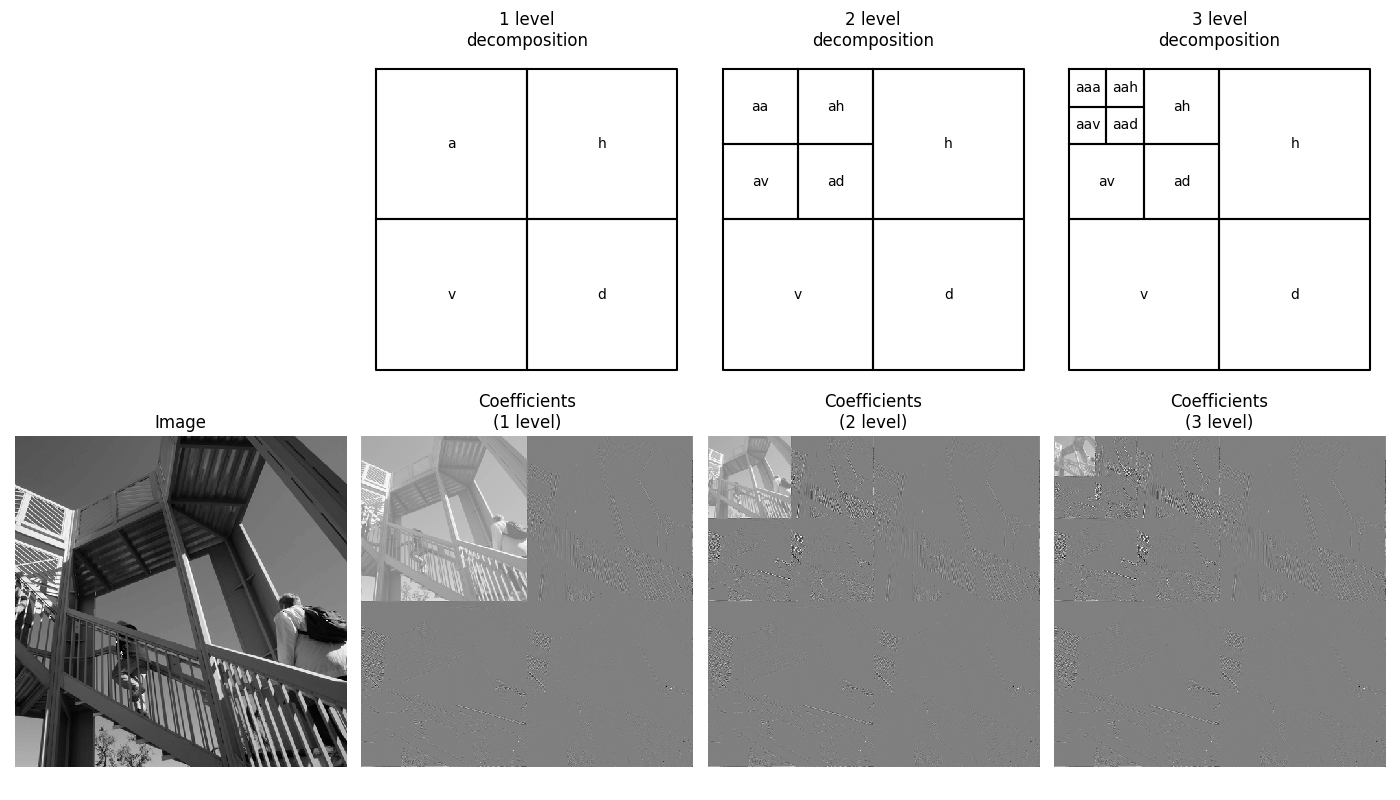}
    \caption{\textbf{The different decomposition levels of the DWT} (e.g., first, second, and third levels): ``a'' indicates the original image, ``h'' indicates the horizontal feature, ``v'' refers to the vertical feature, and ``d'' represents the diagonal feature. The bottom-left image is  the original image, and the following images represent the first-, second-, and third-level decomposition features from the original image.}
    \label{Fig.DWT}
\end{figure}

\begin{figure}
     \centering
     \begin{subfigure}
         \centering
         \includegraphics[width=0.95\textwidth]{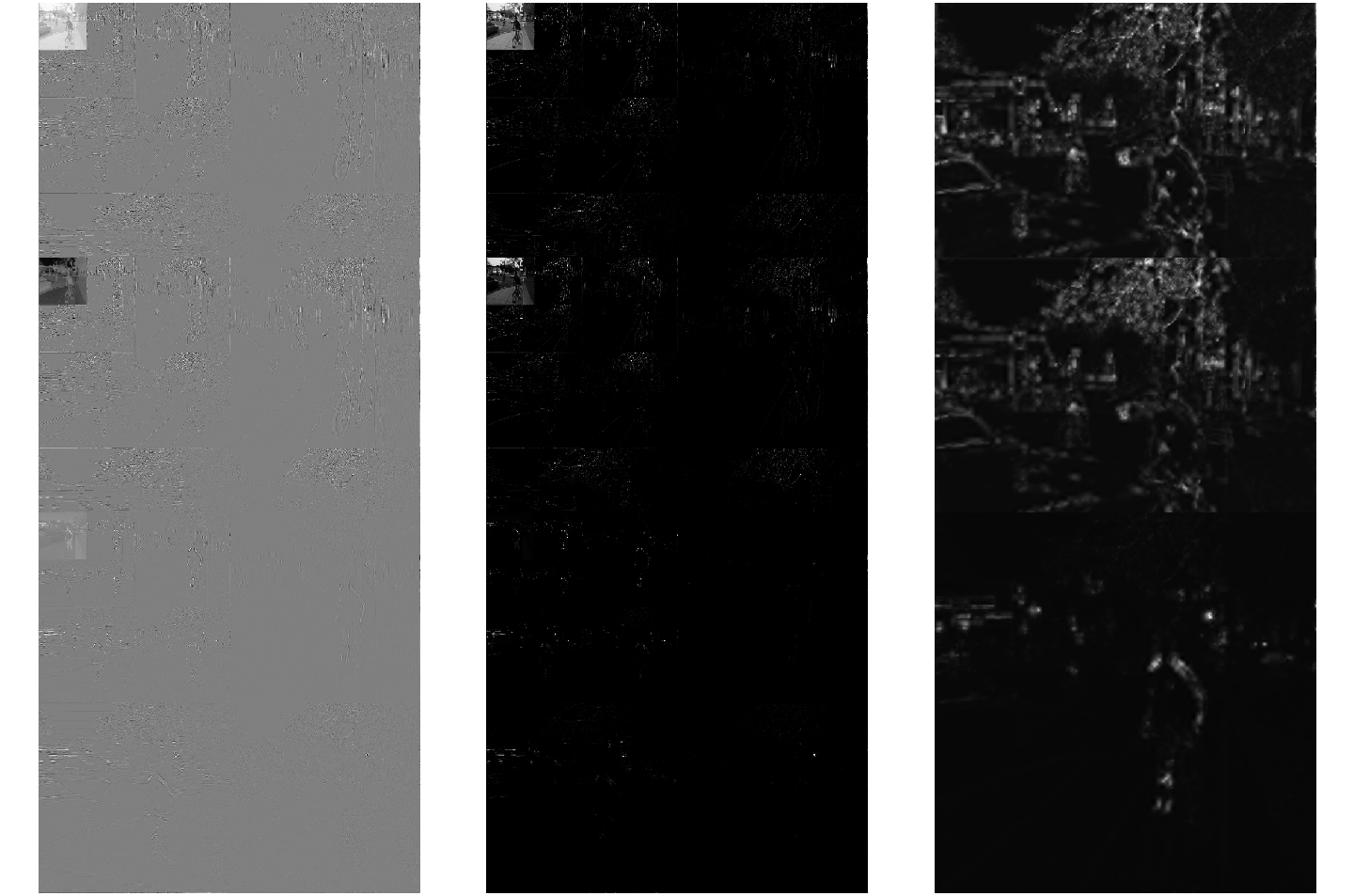}
     \end{subfigure}
     \vfill
     \begin{subfigure}
         \centering
         \includegraphics[width=0.95\textwidth]{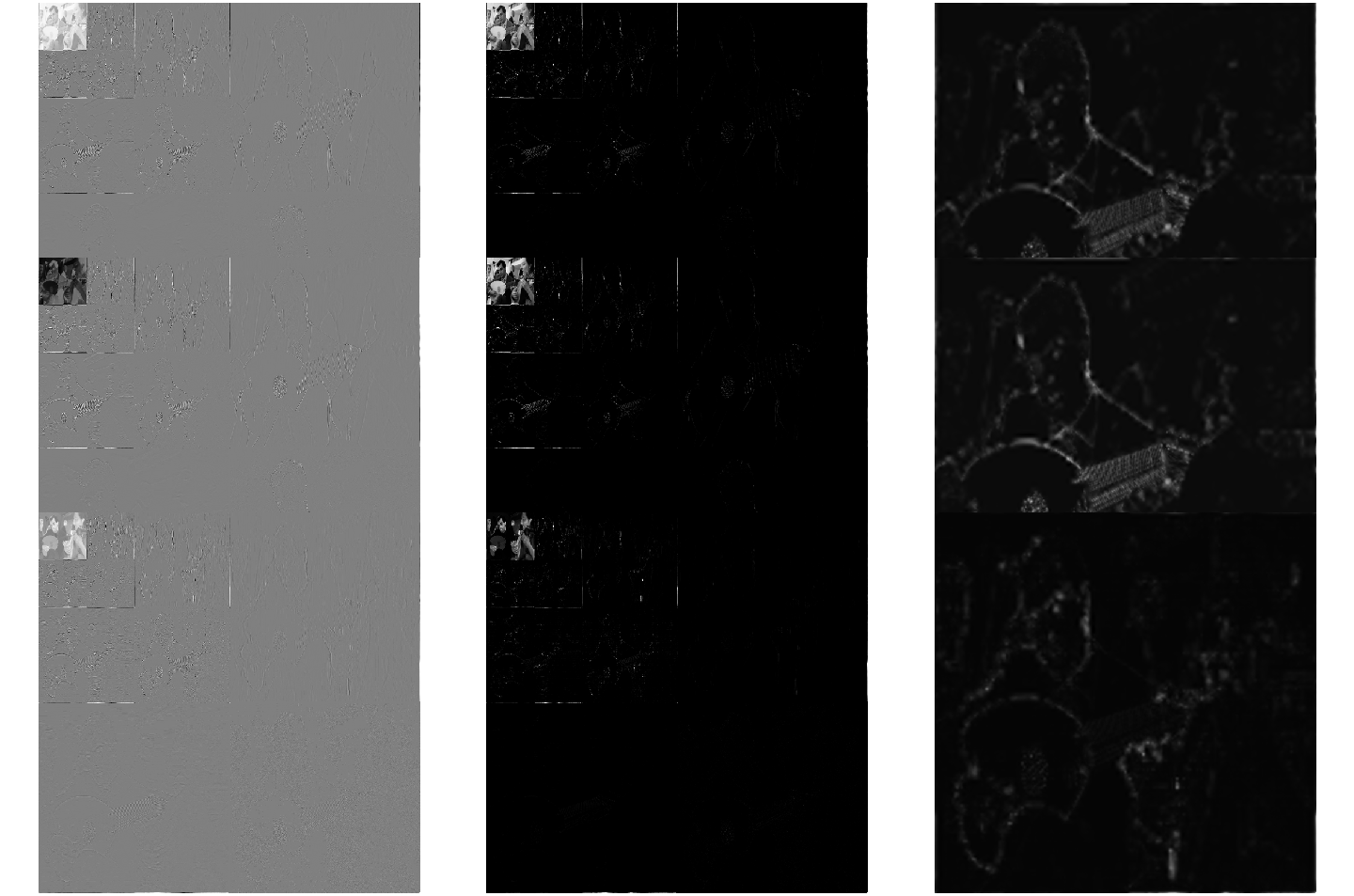}
     \end{subfigure}
     \caption{\textbf{Each channel's DWT map and the wavelet energy maps corresponding to it.} The first column shows the DWT maps for achromatic (WB) and chromatic (RG, YB) channels. The second column is the wavelet energy map, obtained by summing across scales and orientation features for WB, RG, and YB opponent channels, respectively. The last column shows the sum of squares
~energy maps in each opponent channel.}
     \label{Fig.WEP}
\end{figure}

The wavelet image analysis can decompose an image into multi-scale and multi-orientation features, similar to the visual cortex representation. Compared to the Fourier transform (FT), a wavelet transform can represent spatial and frequency information simultaneously. Alfred Haar first proposed the wavelet transform approach, and it has already been widely used in signal analysis~\citep{Haar1912}; for example, for image compression, image denoising, and classification. Wavelet transforms have already been applied in visual saliency map prediction, and achieved good performance~\citep{Nevrez13}. However, wavelet energy maps remain barely used in visual saliency map prediction, and they can be used to enhance local contrast information in the decomposition sub-bands. In our proposed model, we use a discrete wavelet transform (DWT), which can be written in math as: 

\begin{equation}
    r[n]= ((I * f)[n]) \downarrow 2 = (\sum_{k=-\infty}^{\infty} I[k] f[n-k]) \downarrow 2
\end{equation}

where $I$ indicates the input images, $f$ represents a series of filter banks (low-pass and high-pass), and $\downarrow 2$ indicates down-sampling until the next layer's signal cannot be decomposed any more (see Fig.~\ref{Fig.DWT}). A series of sub-band images are produced after convolution with the DWT; then, the wavelet energy map can be calculated from each sub-band feature (see Fig.~\ref{Fig.WEP}).

The wavelet energy map can be expressed as:

\begin{equation}
    \mathcal{WE}(i, j)=\|I(i, j)\|^{2}=\sum_{k=1}^{3ind+1}\left|I_{k}(i, j)\right|^{2},
\end{equation}
 
\textcolor{black}{where $3ind$ indicates the maximum level of an image that can be decomposed in the last level, e.g., 3 level decomposition, and $I_{k}(i, j)^{2}$ represents the energy map of each sub-band feature.}
 
\subsection{Contrast Sensitivity Function}

The human visual system is sensitive to contrast changes
~in natural environments. The visual cortex function can be decomposed into subset compositions, where one of the significant features is the CSF, which can be divided into achromatic and chromatic spatial CSFs~\citep{Mullen85}. In this proposed computational model, an achromatic CSF (aCSF) and chromatic CSFs (rgCSF and ybCSF) were implemented, which was first proposed by Mannos and Sakrison in 1974~\citep{Mannos74}, and further improved later~\citep{Watson02, Watson10} (see Fig.~\ref{Fig.csf}). The achromatic CSF mathematics is as follows: 

\begin{equation}
    \operatorname{CSF}(f_{x}, f_{y})=Q(f) * L(f_{x}, f_{y}),
\end{equation}

\begin{equation}
    Q(f)=g *\left(\exp (-(f/f_{m}))-l* \exp \left(-\left(f^{2} / s^{2}\right)\right)\right),
\end{equation}

\begin{equation}
    L(f_{x}, f_{y})= 1-w *\left(4(1-\exp (-(f/os))) * f_{x}^{2} * f_{y}^{2}\right)/f^{4}),
\end{equation}

where $(f_{x},f_{y})$ indicates a 2D spatial frequency vector (in cycle/deg), $f$ represents the modulus of the spatial frequency (cycle/deg), $g$ represents the overall gain ($g=330.74$), $f_{m}$ is a parameter that controls the exponential decay of the CSF Tyler
~($f_{m}=7.28$), $l$ represents the loss at low frequencies ($l=0.837$), $s$ is a parameter that controls the attenuation of the loss factor at high frequencies ($s=1.809$), $w$ indicates the weighting of the oblique effect ($w=1$), and $os$ indicates the oblique effect scale ($os=6.664$). The CSFs were applied to the wavelet energy image in the Fourier domains. It can be described by the following formula:

\begin{figure}
    \centering
    \includegraphics[width=\textwidth]{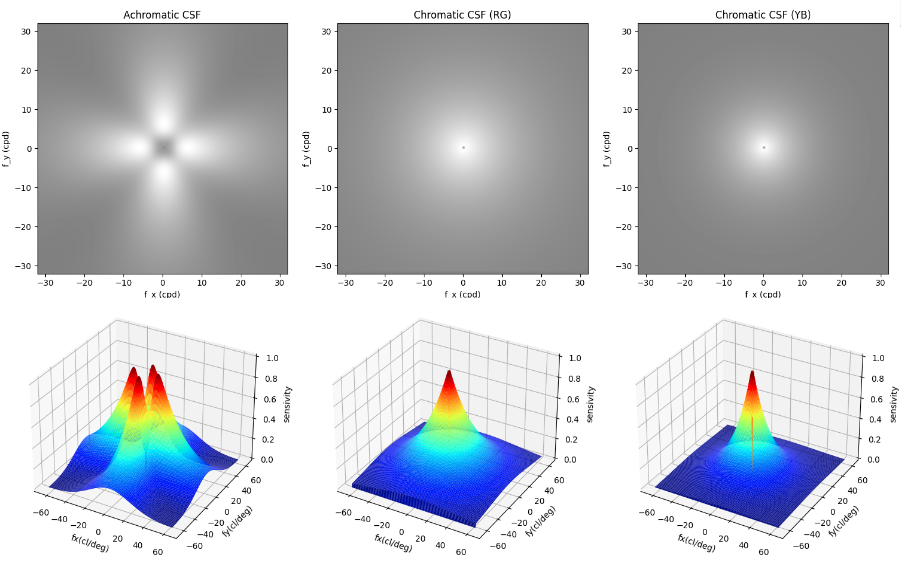}
    \caption{\textbf{Achromatic and chromatic CSFs.} The images in the top row are 2D CSFs, and the bottom row shows 3D CSFs.}
    \label{Fig.csf}
\end{figure}

\begin{equation}
CSF_{WE} = real(\mathcal{F}(\mathcal{I}(\mathrm{I}(\mathrm{F}(WE.real)) \odot CSF))),
\end{equation}

\textcolor{black}{where $\mathrm{F}$ indicates the 2D Fourier transform, $\mathcal{F}$ indicates the 2D inverse Fourier transform, $\mathrm{I}$ indicates \textit{fftshift}, which rearranges a Fourier transform by shifting the zero-frequency component to the center of the image, and $\mathcal{I}$ indicates \textit{ifftshift}, which rearranges a zero-frequency-shifted Fourier transform back to the original transform output. In other words, \textit{ifftshift} undoes the result of \textit{fftshift}.} The Python implementations of the above CSFs (aCSF, rgCSF, and ybCSF) are available at \url{https://github.com/sinodanishspain/CSFpy}.

\section{Materials and Methods}

\subsection{datasets}

The proposed model was tested on several well-known datasets, including MIT1003, MIT300, TORONTO, and SID4VAM. The following sections introduce the basic information of each dataset. 
\begin{itemize}
\item MIT1003 is an image dataset that includes 1003 images from the Flickr
~and LabelMe collections. The fixation map was generated by recording the eye-tracking data of 15 participants. It is the largest eye-tracking dataset. The dataset includes 779 landscape and 228 portrait images with sizes spanning from {$405\times405$ to $1024\times1024$}
~pixels~\citep{Judd12}.

\item MIT300 is a benchmark saliency test dataset that includes 300 images obtained by recoding a 39-observer eye-tracking dataset. The MIT300 dataset categories are highly varied and natural.  The dataset can be used for model evaluation~\citep{Judd12}.

\item TORONTO includes 120 chromatic images free-viewed by 20 subjects. The dataset contains both outdoor and indoor scenes with a fixed resolution of {$511\times681$}
~pixels~\citep{Bruce05}.

\item SID4VAM is a synthetic image database that is mainly used to psychophysically evaluate the V1 properties. This database is composed of 230 synthetic images, including 15 distinct types of low-level features (e.g., brightness, size, color, and orientation) with different target-distractor pop-out-type  synthetic images~\citep{Berga2019SID4VAMAB}.
\end{itemize}

\subsection{Evaluation Metrics}

As mentioned before, there are several approaches to evaluate metrics between visual saliency and model prediction. In general, saliency evaluation can be divided into two branches: location-based and distribution-based. The former mainly focuses on the district located in the saliency map, and the latter considers both the predicted saliency and human eye fixation maps as continuous distributions~\citep{Bylinskii16}. In this research, we used AUC, NSS, CC, SIM, IG, and KL to evaluate the methods and details of each evaluation metric, as described in the following\footnote{https://github.com/cvzoya/saliency}.

\subsubsection{Area under the ROC Curve (AUC)}

The AUC metric is a popular approach for the evaluation of saliency model performance. The saliency map can be treated as a binary classifier to split positive samples from negative samples by setting different thresholds. The true positive (TP) is the proportion of salient map values beyond a specific threshold at the fixation locations. In contrast, the false positive (FP) is the proportion of salient map values beyond a specific threshold at the non-fixation locations. In our case, the thresholds were set from the saliency map values and the AUC-Judd, AUC-Borji, and sAUC measures~\citep{Borji13}.

\subsubsection{Normalized Scanpath Saliency (NSS)}

The NSS metric usually measures the relationship between human eye fixation maps and model-predicted saliency maps~\citep{Emami13}. The NSS can be formally defined as follows given a binary fixation map $F$ and a saliency map $S$: 

\begin{equation}
    N S S=\frac{1}{N} \sum_{i=1}^{N} \bar{S}(i) \times F(i),
\end{equation}

\begin{equation}
    N=\sum_{i} F(i) \text { and } \bar{S}=\frac{S-\mu(S)}{\sigma(S)},
\end{equation}

where $N$ is the total number of human eye positions, $\mu(s)$ is the mean value of saliency maps, and $\sigma(S)$ is the standard deviation.

\subsubsection{Similarity Metric (SIM)}

The similarity metric (SIM) is a very famous algorithm for measuring image structure similarity,
which has already been widely used in image quality and image processing disciplines~\citep{Riche13}. The SIM mainly measures the normalized probability distributions of eye fixation and model-predicted saliency maps. The SIM can be mathematically described as:

\begin{equation}
    SIM=\sum_{i=1} \min (P(i), Q(i)),
\end{equation}

where $P(i)$ and $Q(i)$ are the normalized saliency map and the fixation map, respectively. A similarity score should be in the range between zero and one.

\subsubsection{Information Gain (IG)}

Information gain is an approach to measuring saliency map prediction accuracy from an information-theoretic view. It mainly measures the critical information contained in the predicted saliency map, compared with a ground-truth map~\citep{Kummerer15}. The mathematical formula for the IG can be expressed as:

\begin{equation}
    I G\left(P, Q^{B}\right)=\frac{1}{N} \sum_{i} Q_{i}^{B}\left[\log _{2}\left(\epsilon+P_{i}\right)-\log _{2}\left(\epsilon+B_{i}\right)\right],
\end{equation}

where $P$ indicates the predicted saliency map, $Q^{B}$ is the baseline map, and $\epsilon$ represents a regularity parameter. 

\subsubsection{Pearson’s Correlation Coefficient (CC)}

Pearson’s correlation coefficient (CC) is a linear approach that measures how many similarities there are between the predicted saliency map and the baseline map~\citep{Jost05}.  

\begin{equation}
    C C\left(P, Q^{D}\right)=\frac{\sigma\left(P, Q^{D}\right)}{\sigma(P) \times \sigma\left(Q^{D}\right)},
\end{equation}

where $P$ indicates the predicted saliency map and $Q^{D}$ is the ground-truth saliency map.

\subsubsection{Kullback--Leibler divergence (KL)}

The Kullback-Leibler divergence (KL) is used to measure the distance between the samples of two distributions from an information-theoretic perspective~\citep{Kummerer15}. It can be formally defined as:

\begin{equation}
    KL\left(P, Q^{D}\right)=\sum_{i} Q_{i}^{D} \log \left(\epsilon+\frac{Q_{i}^{D}}{\epsilon+P_{i}}\right),
\end{equation}

where $P$ indicates the predicted saliency map, $Q^{D}$ is the ground-truth saliency map, and $\epsilon$ represents a regularity parameter. 

\subsubsection{Other metrics}

We also evaluated the performance of different salient prediction models through two main metrics: precision-recall curves (PR curves) and the F-measure\footnote{\url{https://github.com/ArcherFMY/sal\_eval\_toolbox/}}. By binarizing the predicted saliency map with thresholds in [0,255], a series of precision and recall score pairs were calculated for each dataset image. The PR curve was plotted using the average precision and recall of the dataset under different thresholds~\citep{Mengyang18}.

\section{Experimental Results}

\subsection{Quantitative Comparison of the Proposed Model with other State-of-the-Art Models}

To evaluate the performance of the proposed model, we compared it with eight other state-of-the-art models. For comparison of the quantitative results, we selected the MIT1003 and SID4VAM benchmarks. These results are reported in Tabs.~\ref{Tab.mit1003}, \ref{Tab.toronto}, and \ref{Tab.sid4vam}. The superior performance, in terms of saliency prediction, was achieved by models based on biological/cognitive and Fourier/spectral foundations. Our model achieved stable and superior performance, in terms of different evaluation metrics compared to other biological/cognitive- and Fourier/spectral-inspired models. However, saliency map prediction based on a convolutional neural network outperformed other models in natural scene images, as more images were used to train the neural network. Consequently, these cannot be compared to other models, as they are based more on statistical than neuroscientific principles. In this paper, we emphasize understanding saliency prediction from a neuroscience perspective, in order to further help us understand the mechanism of visual attention cognitive function. Furthermore, biological/cognitive- and Fourier/spectral-inspired saliency detection models were outperformed by deep learning approaches
~(ML\_Net and DeepGazeII) in the SID4VAM dataset (see Tab.~\ref{Tab.sid4vam} and Fig.~\ref{Fig.SID4VAM_saliencyMap}). As previously said, SID4VAM is a synthetic image database that is primarily used to psychophysically test the V1 properties, which is also why we stated that deep learning models refer more to statistics than neuroscience in the explanation of human visual attention mechanisms.

\begin{table}[!ht]
\centering
\resizebox{0.7\textwidth}{!}{%
\begin{tabular}{|c|c|ccccc|}
\rowcolor[HTML]{00D2CB}
\Xhline{3\arrayrulewidth}
\textbf{Methods} & \textbf{DNN} & \textbf{AUC\_Judd} & \textbf{AUC\_Borji} & \textbf{sAUC} & \textbf{NSS} & \textbf{SIM}  \\
\Xhline{3\arrayrulewidth}
\rowcolor{SkyBlue}
ITT &B       & 0.674 & 0.655 & 0.610  & 0.629  & 0.291 \\
\rowcolor{Green}
WECSF &N        & \textbf{0.705} & \textbf{0.692} & \textbf{0.653} & \textbf{0.849} & \textbf{0.362}\\ 
SR &N          & \textbf{0.708} & 0.683 & 0.638 & 0.791  & 0.329\\
AIM &N         & \textbf{0.706} & \textbf{0.696} & 0.639 & 0.780   & 0.282 \\
BMS &N        & 0.684 & 0.637 & 0.576 & 0.729  & 0.346 \\
CASD &N       & \textbf{0.747} & \textbf{0.731} & 0.651 &\textbf{0.977}  & 0.350  \\
DCTS &N       & \textbf{0.746} & \textbf{0.732} & 0.650  & \textbf{1.000}    & 0.322  \\
HFT &N         & \textbf{0.797} & \textbf{0.764} & 0.619 & \textbf{1.258}  & \textbf{0.416}  \\
ICL &N        & \textbf{0.769} & \textbf{0.713} & 0.617 & \textbf{1.048}  & \textbf{0.420}   \\
PFT &N        & \textbf{0.708} & 0.683 & 0.636 & 0.787  & 0.326 \\
PQFT &N       & 0.643 & 0.530  & 0.519 & 0.459  & 0.292 \\
QDCT &N        & \textbf{0.736} & \textbf{0.714} & 0.647 & \textbf{0.920}   & 0.338\\
RARE &N        & \textbf{0.777} & \textbf{0.755} & \textbf{0.665} & \textbf{1.198}  & \textbf{0.380}\\
SIM &N         & 0.701 & \textbf{0.693} & \textbf{0.653} & 0.743  & 0.283\\
SUN &N        & 0.665 & 0.647 & 0.601 & 0.629  & 0.287 \\
Achanta &N    & 0.534 & 0.526 & 0.526 & 0.174  & 0.240 \\
Simpsal &N     & \textbf{0.735} & \textbf{0.721} & 0.610  & \textbf{0.892}  & 0.337\\
Spratling &N   & 0.512 & 0.508 & 0.510  & 0.039  & 0.234\\
SIMgrouping &N & \textbf{0.724} & \textbf{0.716} & \textbf{0.668} & \textbf{0.873}  & 0.308 \\
SeoMilanfar &N & \textbf{0.710}  & 0.688 & 0.633 & 0.808  & 0.351 \\
\rowcolor{Pink}
\hline
ML\_Net &Y  & \textbf{0.836}   & \textbf{0.743}     & \textbf{0.689}& \textbf{1.928} & \textbf{0.565} \\
\rowcolor{Pink}
DeepGazeII &Y & \textbf{0.886} & \textbf{0.837}   & \textbf{0.779} & \textbf{2.483} & \textbf{0.527}\\
\Xhline{3\arrayrulewidth}
\end{tabular}%
}
\vspace{0.2cm}
\caption{\textbf{Quantitative scores of several models for the MIT1003 dataset.} \textcolor{black}{The baseline ITT model is shaded in light blue and the proposed model is shown in green
. The black-bold scores show that the saliency prediction worked better than our models for certain quantitative metrics. ``N'' indicates NO, ``Y'' indicates YES, and ``B''
~indicates Baseline. The results of  ML\_Net and DeepGazeII models for the MIT1003 dataset  are shown in pink, as this dataset was used to train ML\_Net and DeepGazeII and, so, their results could not be compared with those of the other models.}}
\label{Tab.mit1003}
\end{table}

\begin{table}
\centering
\resizebox{0.7\textwidth}{!}{%
\begin{tabular}{|c|c|ccccc|}
\rowcolor[HTML]{00D2CB} 
\Xhline{3\arrayrulewidth}
\textbf{Methods} &\textbf{DNN} & \textbf{AUC\_Judd} & \textbf{AUC\_Borji} & \textbf{sAUC} & \textbf{NSS} & \textbf{SIM} \\
\Xhline{3\arrayrulewidth}
\rowcolor{SkyBlue}
ITT         &B& 0.700  & 0.679   & 0.641 & 0.816 & 0.317\\ 
\rowcolor{Green}
WECSF       &N& \textbf{0.701} & \textbf{0.686} & \textbf{0.674} & \textbf{0.844}  & \textbf{0.365} \\
SR          &N& \textbf{0.744} & \textbf{0.722} & \textbf{0.683} & \textbf{1.019}  & 0.343 \\
AIM         &N& \textbf{0.727} & \textbf{0.718} & 0.664 & 0.885  & 0.356 \\
SIM         &N& \textbf{0.754} & \textbf{0.744} & \textbf{0.707} & \textbf{0.951}  & 0.361 \\
SUN         &N& 0.674 & 0.653 & 0.613 & 0.656  & 0.285 \\
HFT         &N& \textbf{0.820} & \textbf{0.792} & 0.659 & \textbf{1.548}  & \textbf{0.522} \\
ICL         &N& \textbf{0.792} & \textbf{0.737} & 0.652 & \textbf{1.245}  & \textbf{0.532} \\
PFT         &N& \textbf{0.742} & \textbf{0.717} & \textbf{0.684} & 1.001  & 0.339 \\
CASD        &N& \textbf{0.780} & \textbf{0.764} & \textbf{0.688} & 1.237  & 0.364 \\
PQFT        &N& 0.650 & 0.524 & 0.517 & 0.482  & 0.263 \\
QDCT        &N& \textbf{0.769} & \textbf{0.748} & \textbf{0.691} & 1.174  & 0.354 \\
RARE        &N& \textbf{0.806} & \textbf{0.774} & \textbf{0.693} & 1.514  & \textbf{0.402} \\
Achanta     &N& 0.551 & 0.541 & 0.539 & 0.249  & 0.305 \\
Simpsal     &N& \textbf{0.769} & \textbf{0.754} & 0.648 & \textbf{1.121}  & 0.355 \\
Spratling   &N& 0.508 & 0.503 & 0.509 & 0.015  & 0.242 \\
SIMgrouping &N& \textbf{0.769} & \textbf{0.760} & \textbf{0.710} & \textbf{1.090}   & 0.326 \\
SeoMilanfar &N& \textbf{0.769} & \textbf{0.744} & \textbf{0.695} & \textbf{1.185}  & \textbf{0.382} \\
\rowcolor{Pink}
\hline
ML\_Net     &Y& \textbf{0.823} & \textbf{0.803} & \textbf{0.750} & \textbf{1.824} & \textbf{0.536} \\
\rowcolor{Pink}
DeepGazeII &Y & \textbf{0.846} & \textbf{0.827}  &\textbf{0.756} & \textbf{2.199} & \textbf{0.620}\\
\Xhline{3\arrayrulewidth}
\end{tabular}%
}
\vspace{0.2cm}
\caption{\textbf{Quantitative scores of several models for the TORONTO dataset.} \textcolor{black}{The baseline ITT model is shown in light blue and the proposed model is shown in green. The black-bold scores show that the saliency prediction worked better than our models for certain quantitative metrics. The results of of ML\_Net and DeepGazeII  for the TORONTO dataset are shown in pink, as this dataset was used to train ML\_Net and DeepGazeII and, so, their results could not be compared with those of the other models.}}
\label{Tab.toronto}
\end{table}

\begin{table}
\centering
\resizebox{0.95\textwidth}{!}{%
\begin{tabular}{|c|c|cccccccc|}
\rowcolor[HTML]{00D2CB}
\Xhline{3\arrayrulewidth}
\textbf{Models} & \textbf{DNN} & \textbf{AUC\_Judd} & \textbf{AUC\_Borji} & \textbf{sAUC} &
\textbf{CC} & \textbf{NSS} & \textbf{KL} & \textbf{SIM} & \textbf{IG} \\
\Xhline{3\arrayrulewidth}
\rowcolor{SkyBlue}
GT &B  &0.943 &0.882   &0.860  &1.000 &4.204 &0.000 &1.000  &2.802 \\
\rowcolor{SkyBlue}
Baseline-CG& B &0.703 &0.697  &0.525 &0.281 &1.577&0.722    &0.372 &-0.189\\
\hline
\rowcolor{LightBlue}
ITT&B         & 0.645 & 0.592 & 0.591 & 0.165  & 0.801  & 1.778  & 0.315 & -0.104  \\
\rowcolor{Green}
WECSF&N     & \textbf{0.667} & \textbf{0.647} & \textbf{0.646} & \textbf{0.378}  & \textbf{1.019}  &\textbf{1.769}   & \textbf{0.459} & \textbf{-1.508} \\
AIM&N       & 0.572 & 0.566 & 0.560 & 0.118  & 0.515  & 14.96  & 0.216 & -18.742 \\
Achanta&N     & 0.505 & 0.513 & 0.513 & 0.060  & 0.334  & 8.431  & 0.117 & -9.634  \\
BMS&N         & \textbf{0.811} & 0.517 & 0.515 & 0.129  & 0.702  & 19.161 & \textbf{0.624} & -24.258 \\
CASD&N       & \textbf{0.737} & \textbf{0.673} & \textbf{0.663} & \textbf{0.429}  & \textbf{2.124}  & 2.659  & 0.397 & \textbf{-1.263}  \\
DCTS&N        & \textbf{0.732} & \textbf{0.725} &\textbf{0.715} & \textbf{0.440}  & \textbf{2.191}  &\textbf{1.442}  & 0.377 & \textbf{0.363}   \\
HFT&N         & \textbf{0.780} & \textbf{0.753} &\textbf{0.697} & \textbf{0.561}  & \textbf{2.364}&\textbf{1.424}  & 0.458 & \textbf{0.406}   \\
ICL&N         & \textbf{0.742} & \textbf{0.716} &0.630 & 0.344  & \textbf{1.189}  & 1.974  & 0.391 & \textbf{-0.443}  \\
PFT&N         & \textbf{0.704} & \textbf{0.692} &\textbf{0.688} &\textbf{0.407}  & \textbf{2.074}  & 2.556  & 0.361 & \textbf{-1.230}   \\
PQFT&N       & 0.581 & 0.518 & 0.517 & 0.112  & 0.611  & 12.262 & 0.175 & -15.074 \\
QDCT&N        & \textbf{0.717} & \textbf{0.705} &\textbf{0.699} & \textbf{0.419}  & \textbf{2.139}  & 1.850  & 0.368 & \textbf{-0.213}  \\
SIM&N         & 0.648 & 0.638 &0.628 & 0.181  & 0.744  & 1.799  & 0.340 & -0.152  \\
SR&N          & \textbf{0.747} & \textbf{0.695} &\textbf{0.689} & \textbf{0.423}  & \textbf{2.087}  & \textbf{1.508}  & 0.414 & \textbf{0.354}   \\
SUN&N         & 0.545 & 0.536 & 0.535 & 0.086  & 0.403  & 17.387 & 0.155 & -22.268 \\
Simpsal&N     & \textbf{0.753} & \textbf{0.740} &\textbf{0.661} & 0.370  & \textbf{1.380}   & \textbf{1.461}  & 0.393 & \textbf{0.323}   \\
Spratling&N   & 0.535 & 0.527 & 0.534 & 0.030  & 0.176  & 17.801 & 0.147 & -22.890 \\
SIMgrouping&N & 0.649 & 0.638 & 0.638 & 0.191  & 0.866  & 1.925  & 0.350 & \textbf{-0.338}  \\
SeoMilanfar&N & \textbf{0.703} & \textbf{0.692} &\textbf{0.698} & 0.246  & \textbf{1.181}  & 2.696  & 0.337 & \textbf{-1.400}  \\
\rowcolor{Pink}
\hline
ML\_Net&Y     & \textbf{0.600} & \textbf{0.593} & \textbf{0.501} & \textbf{0.118}  & \textbf{0.341}  & \textbf{7.529}  & \textbf{0.370} & \textbf{-2.153}  \\
\rowcolor{Pink}
DeepGazeII&Y  & \textbf{0.645} & \textbf{0.581} & \textbf{0.475} & \textbf{0.103} & \textbf{0.555} & \textbf{2.450}  & \textbf{0.274} & \textbf{-1.128}  \\ 
\Xhline{3\arrayrulewidth}
\end{tabular}
}
\vspace{0.3cm}
\caption{\textbf{Quantitative scores of several models for the SID4VAM dataset.} \textcolor{black}{The baseline ITT model is shown in light blue and the proposed model is shown in green. The black-bold scores show that the saliency prediction worked better than our models for certain quantitative metrics. The Fourier/spectral-inspired models had the best prediction scores, compared to the other start-of-the-art non-neural network (and even deep neural network) models on the SID4VAM dataset. The results of of ML\_Net and DeepGazeII on the SID4VAM dataset are shown in pink, as this dataset was used to train them, and their results could not be compared with those of the other models
.}}
\label{Tab.sid4vam}
\end{table}

\subsection{Qualitative Comparison of the Proposed Model with Other State-of-the-Art Models}

\begin{figure}[!ht]
    \centering
    \includegraphics[width=\textwidth, height=23em]{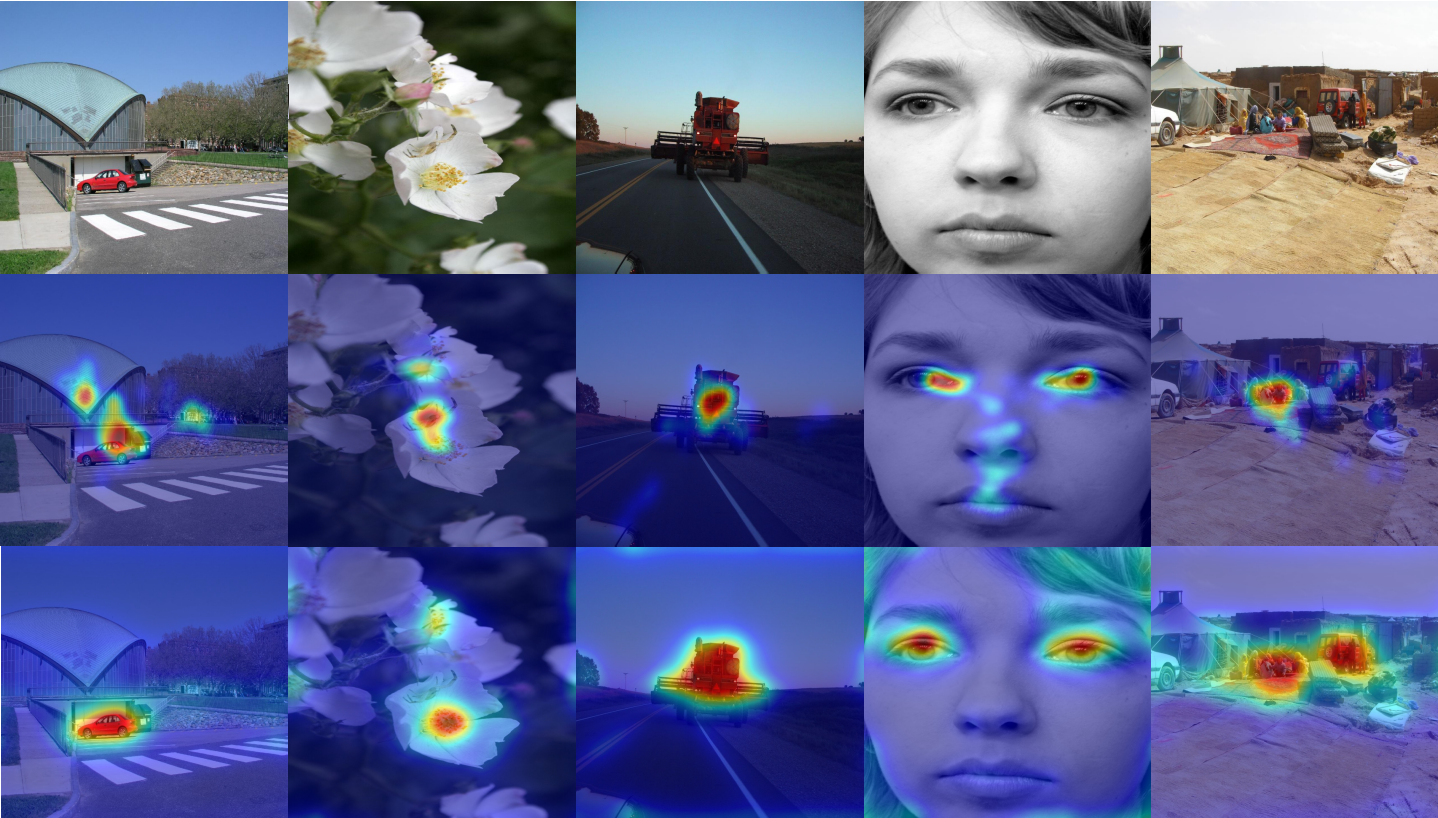}\par 
    \caption{\textbf{Performance evaluation on the MIT1003 dataset.} The first row shows color images, the second row shows ground-truth saliency maps, and the last row shows the proposed model's predicted saliency maps, respectively.}
    \label{Fig.mit1003_saliencyMap}
\end{figure}

\begin{figure}[!ht]
    \centering
    \includegraphics[width=\textwidth, height=20em]{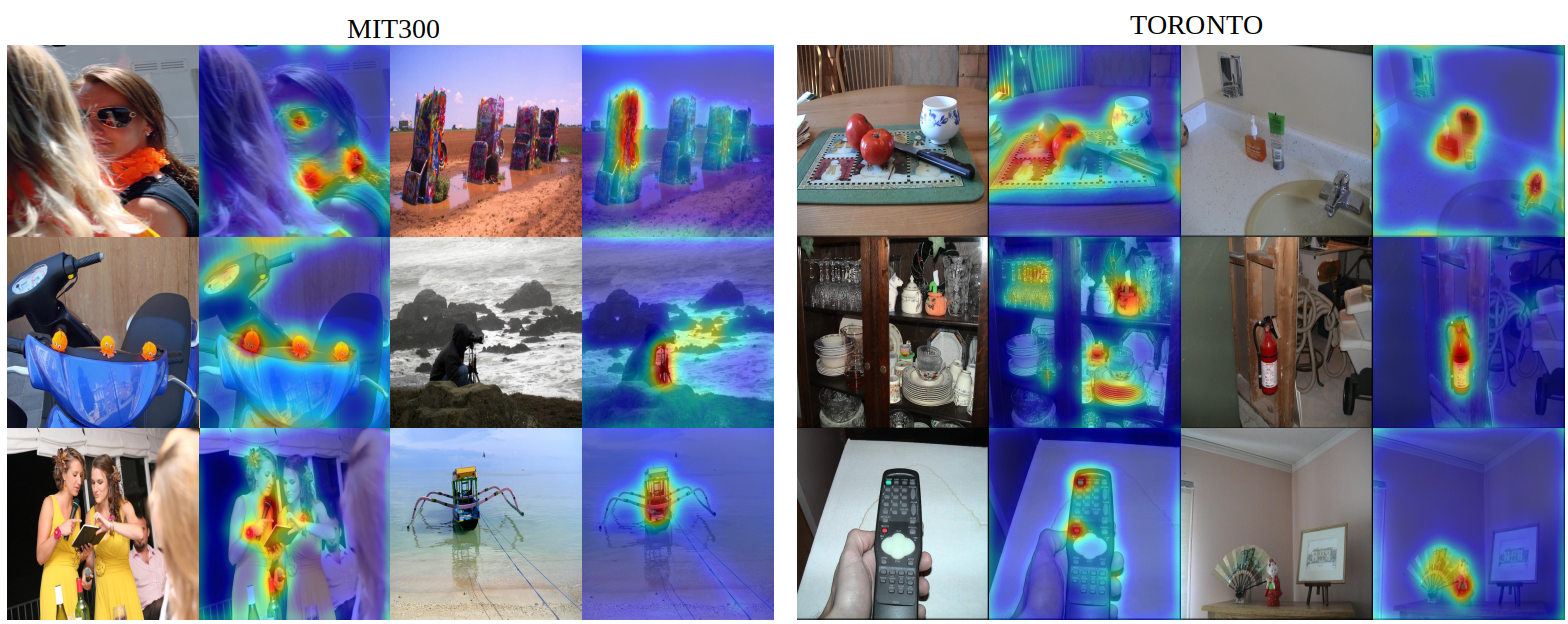}\par 
    \caption{\textbf{Left: Performance evaluation on the MIT300 dataset.} The first and third columns are color images. The second and fourth columns are the  proposed model's predicted saliency maps. \textbf{Right: Performance evaluation on the TORONTO dataset.} The first and third columns are color images. The second and fourth columns are the proposed model's predicted saliency maps.}
    \label{Fig.mit300_saliencyMap}
\end{figure}

\begin{figure}
\centering
\includegraphics[width=0.93\textwidth]{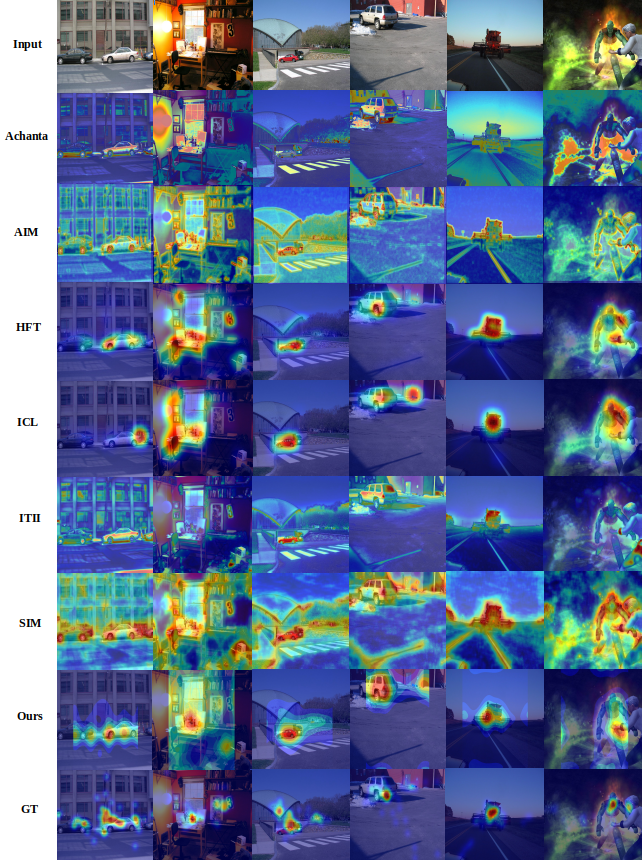}
\caption{\textbf{Qualitative saliency prediction results on the MIT1003 dataset with different models.} The first row shows six stimuli images selected from the MIT1003 dataset. The rows below show the predicted saliency maps obtained with Achanta, AIM, HFT, ICL, ITII, SIM, and the proposed model, as well as the ground-truth (GT) saliency, with artificial color for better visualization.}
\label{Fig.mit1003_gridsaliencyMap}
\end{figure}

\begin{figure}
\centering
\includegraphics[width=0.98\textwidth]{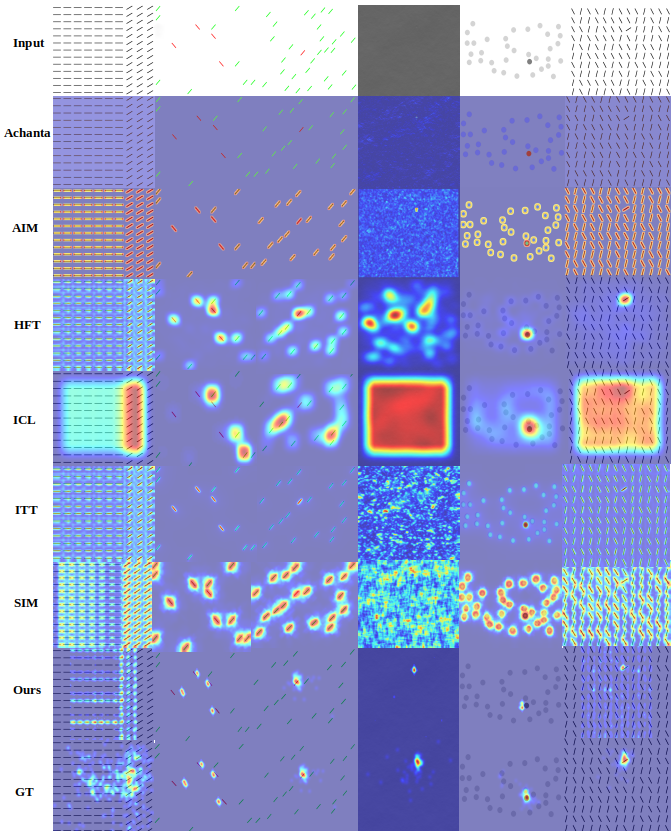}
\caption{\textbf{Qualitative saliency prediction results on the SID4VAM dataset with different models.} The first row shows six stimuli images selected from the SID4VAM dataset. The rows beneath show the salience prediction results obtained with Achanta, AIM, HFT, ICL, ITII, SIM, and the proposed model, as well as the ground truth (GT) salience,
~with artificial color for better visualization. The proposed model can be successfully applied to explain the ``pop-out'' effects in the visual search.}
\label{Fig.SID4VAM_saliencyMap}
\end{figure}

\begin{center}
    \begin{figure}[!ht]
    \centering
    \includegraphics[width=0.95\textwidth]{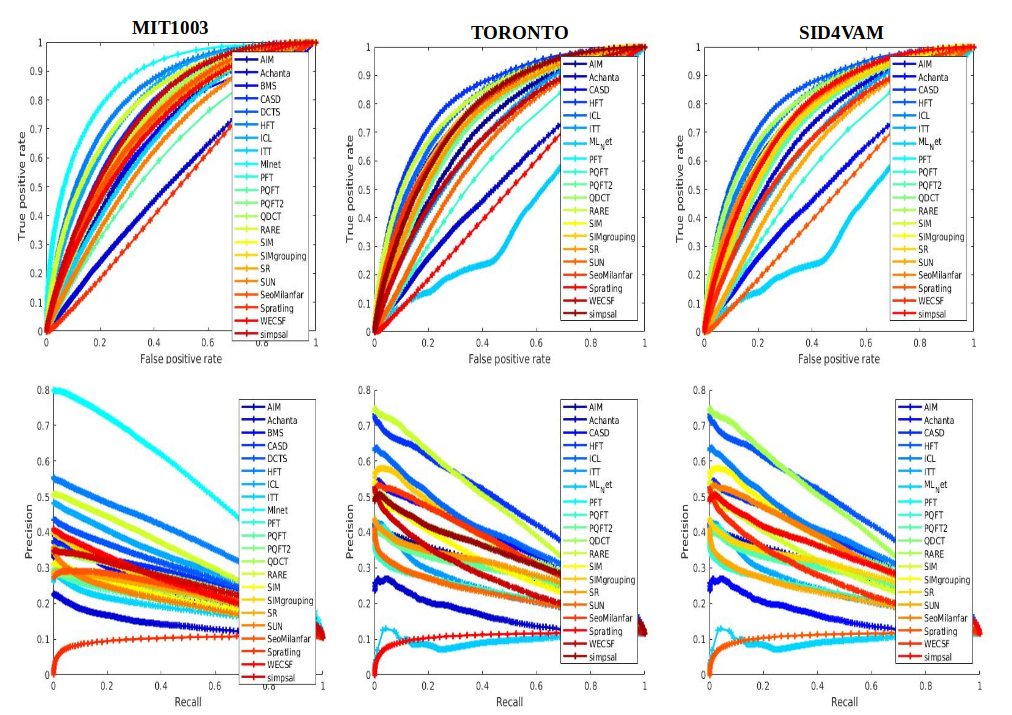}
    \caption{\textbf{ROC curve (AUC) and PR curves.} Comparison of the area under the ROC curve (AUC) and PR curves, with different thresholds between our method and other state-of-the-art methods on three benchmark datasets.}
    \label{Fig.auc_pr}
    \end{figure}
\end{center}

\begin{figure}
    \centering
    \includegraphics[width=\textwidth]{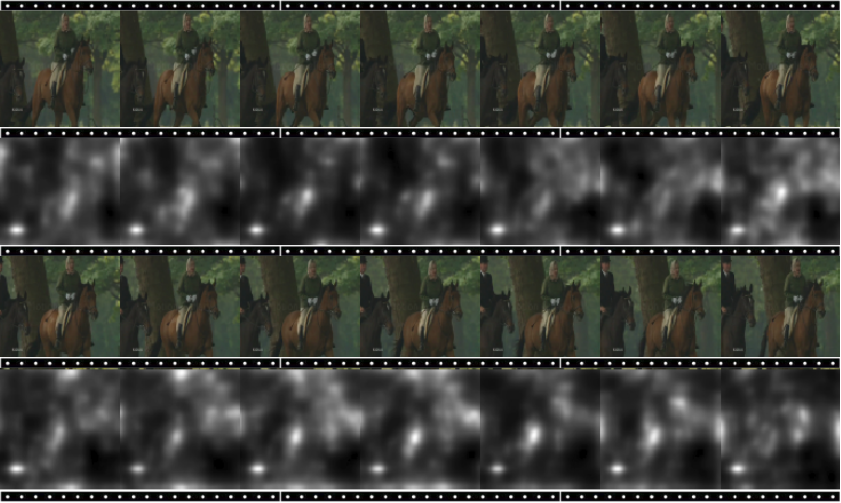}
    \caption{\textbf{Dynamic saliency prediction.} For these sample frames from the UCF Sports Action dataset, the model clearly produced better results and perfectly captured the text information on the bottom-left.}
    \label{Fig.ucsf_saliencyMap}
\end{figure}

We qualitatively tested the proposed model using the MIT1003, MIT300, TORONTO, SID4VAM, and UCF Sports datasets\footnote{\url{https://www.crcv.ucf.edu/data/UCF\_Sports\_Action.php}}. We also compared the model's performance with that of other state-of-the-art saliency prediction models on the MIT1003, TORONTO, and SID4VAM datasets. Figs.~\ref{Fig.mit1003_saliencyMap}, \ref{Fig.mit300_saliencyMap}, \ref{Fig.mit1003_gridsaliencyMap}, \ref{Fig.SID4VAM_saliencyMap}, and \ref{Fig.ucsf_saliencyMap} show the saliency map results when the proposed model and other state-of-the-art models were applied to sample images from the studied datasets. The performance each of saliency prediction model was evaluated through AUC and PR curves, as shown in Fig.~\ref{Fig.auc_pr}. We can see that the proposed model could predict most of the salient objects in the given images. Furthermore, the proposed model could successfully detect the orientation, boundary, and pop-out functions when the model was applied to the SID4VAM dataset. In summary, our proposed biological/Fourier/spectral inspired-saliency prediction model achieved superior and stable performance on natural images, psychophysical synthetic images, and dynamic scenes, compared with other existing models. 

\section{Discussion}

Saliency modeling has become a well-known area of study in computer vision and neuroscience.As a result, academics have attempted to employ different architectures that have not significantly increased model performance. Instead, we should look into how humans perceive scenes; what draws their attention is crucial. In this study, we addressed numerous methods for going beyond the capabilities of such models.

First, it is vital to comprehend the cognitive attention mechanism. Visual attention is a selective cognitive process that helps us deal with this issue successfully by focusing on key information while disregarding unnecessary information. Spatial attention is important in discrimination and appearance tasks in saliency prediction studies. Investigating the brain underpinnings of visual attention can help us design better and more accurate saliency prediction models. On the other hand, better saliency prediction models can assist us in comprehending the cognitive process of visual attention in the brain. Second, we require multi-model and multi-label datasets to assess the effectiveness of saliency prediction models. The MIT Saliency benchmark datasets were used to evaluate the saliency prediction model. As previously stated, the saliency prediction model should perform better on natural scene photos and psychophysical synthetic images (e.g., SID4VAM). This could aid in the improvement of the model architecture and our understanding of the cognitive attention process. Third, we offered extensive experimental findings demonstrating that our method consistently achieved steady and better results when compared to other state-of-the-art technologies. It's worth noting that we used biologically inspired visual model estimation to determine saliency. Our proposed saliency prediction model incorporates more neuroscience notions than statistical concepts.

Furthermore, the suggested model contains a small number of parameters that were essentially set and fixed for all tests. One of the most important things that contributed to the suggested method's efficiency was the use of wavelet energy maps and opponent CSFs as features. Furthermore, our extensive experimental results showed that the proposed saliency prediction measure generated from a local image energy estimator is far more successful and straightforward to implement than existing methods. Even though our method was built solely on biologically inspired computational principles, the resulting model structure showed significant agreement with the fixation behavior of the human visual system.

\section{Conclusions}

In this study, a computational psychophysical visual saliency prediction model inspired by a low-level human visual pathway was proposed. The model includes color opponent channels, wavelet transform, wavelet energy map, and contrast sensitivity functions in order to predict the saliency map. The model was evaluated by classical benchmark datasets and achieved strongly stable and better performance in terms of visual saliency prediction, compared with the baseline model. Furthermore, we found that models based on deep neural networks outperformed ours in terms of natural image salience prediction but underperformed for psychophysical synthetic images. In contrast, Fourier/spectral-inspired models had the opposite effect, as Fourier/spectral-inspired models simulate optical neural processing from the retina to the V1. However, deep neural networks take statistics into account more than low-level vision system functioning, and we argue that deep neural networks can't reliably predict salient performance on psychophysical synthetic images without using specialized architectures and goals. Lastly, we added spatial-temporal saliency prediction to our model, and it was able to pick out the most important thing in the videos.

\section{Data and Code availability}

The code performs main part of experiments described in this article are available at project page: \url{https://sinodanishspain.github.io/HVS_SaliencyModel/}.

\section{Acknowledgments}

I thank the anonymous reviewers, whose suggestions helped to improve and clarify this manuscript. This work was partially funded by these spanish/european grants from GVA/AEI/FEDER/EU: MICINN PID2020-118071GB-I00, MICINNPDC2021-121522-C21, and GVA Grisolía-P/2019/035.

\section{Conflicts of Interest}
The author declares
~no conflicts of interest.

\section{Abbreviations}
\noindent
\begin{tabular}{@{}ll}
HVS & Human Vision System \\
V1 & Primary Visual Cortex\\
ICL & Incremental Coding Length\\
CNN & Convolution Neural Network \\
DNN & Deep Neural Network \\
WT & Wavelet Transform \\
IWT & Inverse Wavelet Transform\\
AUC &  Area Under the ROC Curve  \\
NSS & Normalized Scanpath Saliency \\
CC & Pearson’s Correlation Coefficient \\ 
SIM & Similarity or Histogram Intersection \\
IG & Information Gain  \\
KL &  Kullback--Leibler divergence\\ 
CSF & Contrast Sensitivity Function \\
FT & Fourier Transform \\
DWT & Discrete Wavelet Transform \\
IDWT & Inverse  Discrete Wavelet Transform \\
LGN & Lateral Geniculate Nucleus \\
GT & Ground Truth 
\end{tabular}

\bibliographystyle{abbrvnat}
\bibliography{main}

\end{document}